\documentclass[10pt,twocolumn,letterpaper]{article}

\usepackage[pagenumbers]{cvpr} 

\newcommand{\nop}[1]{}

\DeclareMathOperator{\encoder}{Encoder}
\DeclareMathOperator{\decoder}{Decoder}
\DeclareMathOperator{\head}{Head}
\DeclareMathOperator{\memory}{MemFormer}
\DeclareMathOperator{\context}{CtxFormer}
\DeclareMathOperator{\mlp}{MLP}
\DeclareMathOperator{\match}{IsMatch}
\DeclareMathOperator{\distance}{Dist}

\usepackage{graphicx}
\usepackage{amsmath}
\usepackage{amssymb}
\usepackage{booktabs}
\usepackage{multirow}
\usepackage{makecell}
\usepackage{algorithm}
\usepackage{algorithmic}
\usepackage{mathtools}
\usepackage{overpic}
\usepackage{url}
\usepackage{color}
\usepackage{colortbl}
\usepackage{xcolor}
\usepackage{pifont}
\usepackage{epstopdf}
\usepackage{svg}
\usepackage{caption}
\usepackage{subcaption}
\usepackage{bm}
\usepackage{dsfont}

\definecolor{cvprblue}{rgb}{0.21,0.49,0.74}
\usepackage[pagebackref,breaklinks,colorlinks,allcolors=cvprblue]{hyperref}

\def\paperID{4762} 
\def\confName{CVPR}
\def\confYear{2025}

\title{ModeSeq: Taming Sparse Multimodal Motion Prediction with Sequential Mode Modeling}

\author{
Zikang Zhou$^{1}$~~~
Hengjian Zhou$^{2}$~~~
Haibo Hu$^{1}$~~~
Zihao Wen$^{1}$~~~\\
Jianping Wang$^{1}$~~~
Yung-Hui Li$^3$~~~
Yu-Kai Huang$^4$~~~
\\
$^1$City University of Hong Kong~~~\\
$^2$Zhejiang University~~~
$^3$Hon Hai Research Institute~~~
$^4$Carnegie Mellon University~~~
}

\begin{document}
\maketitle
\begin{abstract}
Anticipating the multimodality of future events lays the foundation for safe autonomous driving. However, multimodal motion prediction for traffic agents has been clouded by the lack of multimodal ground truth. Existing works predominantly adopt the winner-take-all training strategy to tackle this challenge, yet still suffer from limited trajectory diversity and uncalibrated mode confidence. While some approaches address these limitations by generating excessive trajectory candidates, they necessitate a post-processing stage to identify the most representative modes, a process lacking universal principles and compromising trajectory accuracy. We are thus motivated to introduce ModeSeq, a new multimodal prediction paradigm that models modes as sequences. Unlike the common practice of decoding multiple plausible trajectories in one shot, ModeSeq requires motion decoders to infer the next mode step by step, thereby more explicitly capturing the correlation between modes and significantly enhancing the ability to reason about multimodality. Leveraging the inductive bias of sequential mode prediction, we also propose the Early-Match-Take-All (EMTA) training strategy to diversify the trajectories further. Without relying on dense mode prediction or heuristic post-processing, ModeSeq considerably improves the diversity of multimodal output while attaining satisfactory trajectory accuracy, resulting in balanced performance on motion prediction benchmarks. Moreover, ModeSeq naturally emerges with the capability of mode extrapolation, which supports forecasting more behavior modes when the future is highly uncertain.
\end{abstract}    
\section{Introduction}
\label{sec:intro}

\begin{figure}
     \centering
     \begin{subfigure}[b]{0.47\textwidth}
         \centering
         \includegraphics[width=\textwidth]{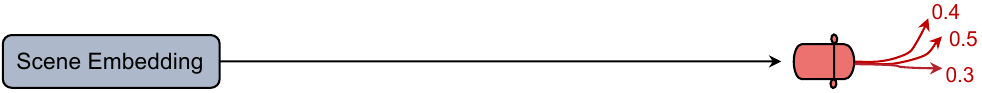}
         \caption{Parallel mode modeling}
         \label{fig:parallel_mode}
     \end{subfigure}
     \hspace{\textwidth}

     \begin{subfigure}[b]{0.47\textwidth}
         \centering
         \includegraphics[width=\textwidth]{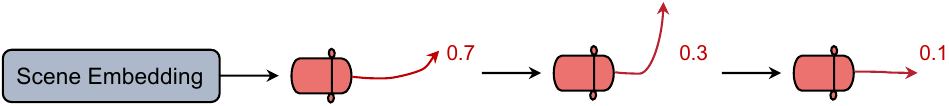}
         \caption{Sequential mode modeling}
         \label{fig:sequential_mode}
     \end{subfigure}
        \vspace{-0.3cm}
        \caption{A comparison between parallel and sequential mode modeling. While parallel mode modeling (\cref{fig:parallel_mode}) decodes multimodal trajectories in one shot, our sequential mode modeling (\cref{fig:sequential_mode}) reasons about multiple plausible futures step by step, which captures the relationships between modes to avoid producing indistinguishable trajectories and confidence scores.}
        \label{fig:comparison}
        \vspace{-0.8cm}
\end{figure}

Handling the intricate uncertainty presented in the real world is one of the major hurdles in autonomous driving. One aspect of the uncertainty lies in the multimodal behavior of traffic agents, \ie, multiple instantiations of an agent's future may be compatible with a given observation of the past. Without characterizing the multimodal distribution of agent motions, autonomous vehicles may fail to interact with the surroundings in a safe and human-like manner. For this reason, advanced decision-making systems demand a motion predictor to forecast several plausible and representative trajectories of critical agents~\cite{ding2021epsilon, li2023marc}.

Although multimodality has long been the central topic studied in motion prediction, this problem has not been fundamentally solved owing to the unavailability of multimodal ground truth, \ie, only one possibility is observable in real-world driving data. To struggle with this dilemma, most existing works adopt the winner-take-all (WTA) strategy~\cite{lee2016stochastic} for training~\cite{cui2019multimodal, chai2019multipath, liang2020learning, zhou2022hivt, varadarajan2022multipath++, nayakanti2022wayformer, zhou2023query, shi2022mtr}. Under this strategy, only the best among all predicted modes will receive explicit supervision signals for trajectory regression and confidence scoring, while all the remaining will be masked in loss calculation. Despite being the current standard practice in the research community, the WTA solution has been found to cause mode collapse easily and produce indistinguishable trajectories~\cite{makansi2019overcoming, rupprecht2017learning, thiede2019analyzing}, further confusing the learning of mode scoring~\cite{lin2024eda}. As a remedy, some recent research intends to cover the ground-truth mode by generating a massive number of trajectory candidates~\cite{varadarajan2022multipath++, nayakanti2022wayformer, shi2022mtr}, from which the most representative ones are heuristically selected based on post-processing methods such as non-maximum suppression (NMS). However, such a post-processing step requires carefully tuning the hyperparameters, \eg, the thresholds in NMS. Even if the hyperparameters were well-tuned, they might not fit various scenarios with diverse road conditions, leading to inferior generalization. Moreover, performing dense mode prediction followed by rule-based post-processing may significantly sacrifice trajectory accuracy in practice~\cite{shi2022mtr}, given that correctly extracting the best trajectories from a large set of candidates is non-trivial.

The limitations of mainstream methods prompt us to seek an end-to-end solution that directly produces a sparse set of diverse, high-quality, and representative agent trajectories, eliminating the need for dense mode prediction and heuristic mode selection. To begin with, we identify a commonality of existing multimodal motion predictors, that all trajectory modes are decoded in one shot, which we dub parallel mode modeling as depicted in \cref{fig:parallel_mode}. Despite its efficiency, this paradigm neglects the relationship between the predicted modes, hindering models from decoding diverse multimodal output. For anchor-free methods with parallel mode modeling~\cite{ngiam2021scene, varadarajan2022multipath++, zhou2022hivt, nayakanti2022wayformer, zhou2023query}, the distinction between the decoded trajectories depends entirely on the difference in sub-network parameters, which may not be controllable under the unstable WTA training. For this reason, some of these solutions turn to dense mode prediction and rely on post-processing steps to diversify the output~\cite{varadarajan2022multipath++, nayakanti2022wayformer}. While anchor-based approaches offload the duties of ensuring diversity onto the input anchors~\cite{phan2020covernet, chai2019multipath, zhao2020tnt, shi2022mtr}, determining a sparse set of anchors that can adapt to specific scenarios is challenging, compelling all these approaches to employ excessive anchors for dense mode prediction. Under the paradigm of parallel mode modeling, producing multimodal trajectories with sparse mode prediction faces significant obstacles.

To tackle these challenges, this paper explores sequential mode modeling (\textbf{ModeSeq}), a completely different pathway toward sparse multimodal motion prediction. As illustrated in \cref{fig:sequential_mode}, we attempt to construct a chain of modes when decoding the future from the scene embeddings, producing only one plausible trajectory and the corresponding confidence at each decoding step. Compared with parallel prediction, our approach puts more effort into capturing the correlation between modes, asking the model to tell what the next mode should be and how much confidence it has conditioned on the mode embeddings at previous steps. By giving the model a chance to look at the prior modes and learning the factorized joint latent space of multiple futures, we tremendously boost the capability of reasoning about multimodality and characterizing the full distribution without the reliance on dense mode prediction, post-processing tricks, and all manner of anchors. Leveraging the order of modes in the sequence, we further propose the Early-Match-Take-All (EMTA) training strategy, which facilitates decoding matched trajectories with high confidence as early as possible. Meanwhile, our EMTA scheme enforces the decoder to vacate the duplicated modes to cover the missing futures at the cost of negligible degradation in trajectory accuracy, thereby achieving better mode coverage and easing the learning of confidence scoring. Furthermore, we strengthen the model capacity under our new paradigm by developing an iterative refinement framework similar to DETR-like decoders~\cite{carion2020end, zhou2023query, shi2022mtr, zhou2024smartrefine}, which integrates a mode rearrangement mechanism in between decoding layers to coordinate with the EMTA training scheme. 

We validate ModeSeq on the Waymo Open Motion Dataset~\cite{ettinger2021large} and the Argoverse 2 Motion Forecasting Dataset~\cite{Argoverse2}, where we achieve more balanced performance in terms of mode coverage, mode scoring, and trajectory accuracy, compared with representative motion forecasting methods such as QCNet~\cite{zhou2023query} and MTR~\cite{shi2022mtr}. Furthermore, our approach naturally emerges with the capability of mode extrapolation thanks to sequential mode modeling, which enables predicting a dynamic number of modes on demand.

\section{Related Work}

\noindent \textbf{Multimodality} has been a dark cloud in the field of motion prediction. Early works employ generative models to sample multimodal trajectories~\cite{tang2019multiple, rhinehart2019precog, lee2017desire, gupta2018social, hong2019rules}, but they are susceptible to mode collapse. Modern motion predictors~\cite{cui2019multimodal, chai2019multipath, liang2020learning, zhou2022hivt, varadarajan2022multipath++, nayakanti2022wayformer, zhou2023query, shi2022mtr} mostly follow the paradigm of multiple choice learning~\cite{lee2016stochastic}, where multiple trajectory modes are produced directly from mixture density networks~\cite{bishop1994mixture}. Due to the lack of multimodal ground truth, these methods adopt the WTA training strategy~\cite{lee2016stochastic}, which is unstable and fails to deal with mode collapse fundamentally~\cite{makansi2019overcoming, rupprecht2017learning, thiede2019analyzing}. To mitigate this issue, a line of research performs dense mode prediction, \ie, decoding excessive trajectory candidates for better mode coverage~\cite{varadarajan2022multipath++, nayakanti2022wayformer, shi2022mtr}. Among these works, some equip the decoder with anchors~\cite{shi2022mtr, zhao2020tnt, chai2019multipath, phan2020covernet} to achieve more stable training. However, dense mode prediction necessitates heuristic selection of the most representative trajectories from a large set of candidates during inference, risking the precision of multimodal output and the generalization across a wide range of scenarios. This paper provides new insights into multimodal problems by introducing the paradigm of sequential mode modeling and the EMTA training strategy, pursuing an end-to-end solution that produces a sparse set of diverse, high-quality, and representative trajectories directly.
\noindent \textbf{Sequential modeling} has found many applications in motion prediction and traffic modeling. On the one hand, applying sequential modeling to the time dimension results in trajectory encoders and decoders based on recurrent networks~\cite{hochreiter1997long, cho2014learning, alahi2016social, deo2018convolutional, rhinehart2019precog, rhinehart2018r2p2, mercat2020multi, lee2017desire, gupta2018social, salzmann2020trajectron++, varadarajan2022multipath++, tang2019multiple} or Transformers~\cite{vaswani2017attention, ngiam2021scene, liu2021multimodal, giuliari2021transformer, yu2020spatio, nayakanti2022wayformer, zhou2022hivt, zhou2023query, seff2023motionlm, zhou2024behaviorgpt, philion2024trajeglish}, which can facilitate the learning of temporal dynamics. In particular, recent advances in motion generation~\cite{seff2023motionlm, philion2024trajeglish, zhou2024behaviorgpt} have shown that factorizing the joint distribution of multi-agent time-series in a social autoregressive manner~\cite{alahi2016social, tang2019multiple, rhinehart2019precog} can better characterize the evolution of traffic scenarios. On the other hand, some works utilize sequential modeling in the agent dimension for multi-agent motion prediction~\cite{sun2022m2i, rowe2023fjmp}. For example, M2I~\cite{sun2022m2i} uses heuristic methods to label influencers and reactors from pairs of agents, followed by predicting the marginal distribution of the influencers and the conditional distribution of the reactors. FJMP~\cite{rowe2023fjmp} extends M2I to model the joint distribution of an arbitrary number of agents, where the joint future trajectories of agents are factorized using a directed acyclic graph. Our work is the first attempt that employs sequential prediction in the mode dimension, which enhances the understanding of multimodal behavior by capturing the correlation between modes.
\section{Methodology}

\begin{figure*}
  \centering
  \includegraphics[width=1.0\linewidth]{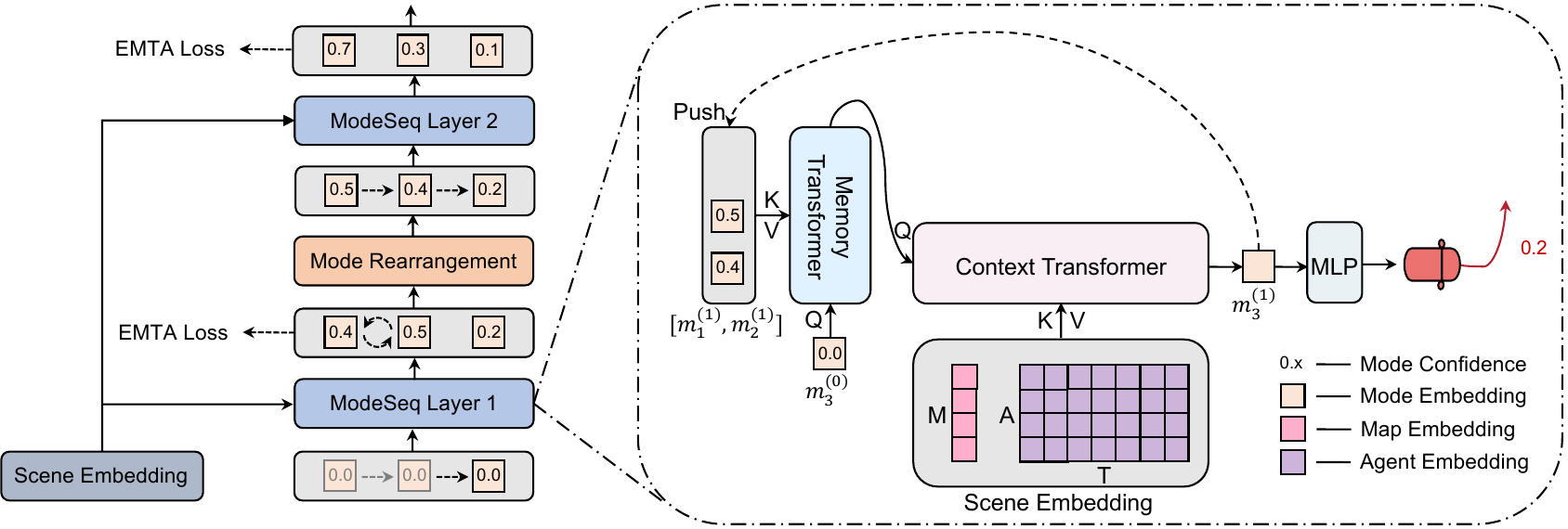}
  \vspace{-0.7cm}
      \caption{Overview of the ModeSeq framework. \textbf{Left:} We stack multiple ModeSeq layers with mode rearrangement in between to iteratively refine the multimodal output under the Early-Match-Take-All (EMTA) training strategy. \textbf{Right:} Each ModeSeq layer consists of a Memory Transformer module for capturing mode-wise dependencies and a Context Transformer module for retrieving the scene embeddings produced by the encoder, operating in a recurrent fashion to decode a sequence of trajectory modes.}
  \label{fig:modeseq}
  \vspace{-0.6cm}
\end{figure*}

\subsection{Problem Formulation}

Denote $S$ as the input of motion prediction models, which encompasses the map elements represented as $M$ polygonal instances and the $T$-step historical trajectories of $A$ traffic agents (\eg, vehicles, pedestrians, and cyclists) in the scene. Each map polygon comprises line segments outlining its contour and semantic tags describing the traffic rules it associates. On the other hand, each trajectory is labeled with agent type (\eg, vehicles, pedestrians, and cyclists) and composed of 3D bounding boxes across various time steps. The models are tasked with forecasting $K$ plausible trajectory modes per agent of interest, each comprising $\hat{T}$ waypoints and an associated confidence score. The $K$ confidence scores are not restricted to a sum of $1$, as it is sufficient to only care about their relative ranking. The predicted trajectories are desired to be representative, reflecting distinct behavior modes of agents and properly measuring the likelihood of each mode via the estimated confidence.

Typical motion predictors employ the encoder-decoder architecture, where an encoder computes the embedding $\Psi$ from the scene input, based on which a decoder learns the embeddings $\{\bm{m}_{i,k}\}_{k \in \{1,\ldots,K\}}$ of the $i$-th agent's future modes. Without loss of generality, the following simplifies $\bm{m}_{i,k}$ as $\bm{m}_k$ to discuss the prediction for a single agent, which can be extended for multiple agents by repeating the same decoding process. Given $\bm{m}_k$, a prediction head then outputs a trajectory $\hat{\bm{y}}_k = [\hat{\bm{y}}_k^1,\ldots,\hat{\bm{y}}_k^{\hat{T}}]$ and a confidence score $\hat{\bm{\phi}}_k$ via simple modules such as multi-layer perceptrons (MLPs). The whole pipeline can be summarized as
\vspace{-0.2cm}
\begin{equation}
    \begin{dcases}
        \Psi = \encoder\left(S\right) \,, \\
        \{\bm{m}_k\}_{k \in \{1,\ldots,K\}} = \decoder\left(\Psi\right) \,, \\
        \hat{\bm{y}}_k, \hat{\bm{\phi}}_k = \head\left(\bm{m}_k\right), \quad k \in \{1,\ldots,K\} \,. \\
    \end{dcases}
\end{equation}

\subsection{Scene Encoding}
\label{sec:encoder}

Since this work focuses on the decoding of multimodal trajectories, we mainly adopt QCNet~\cite{zhou2023query} as the scene encoder, which is one of the \textit{de facto} best practices in industry and academia due to its symmetric modeling in space and time leveraging relative positional embeddings~\cite{shaw2018self}. This encoder, on the one hand, exploits a hierarchical map encoding module based on map-map self-attention to produce the map embedding of shape $[M, D]$, with $D$ referring to the hidden size. On the other hand, the encoder consists of Transformer modules that factorize the space and time axes, including temporal self-attention, agent-map cross-attention, and agent-agent self-attention. These three types of attention are grouped and interleaved twice to yield the agent embedding of shape $[A, T, D]$ for $A$ agents and $T$ historical time steps, which constitutes the final scene embeddings together with the map embedding. In principle, any scene encoding method can fit into our ModeSeq framework with reasonable efforts (see the supplementary material).

\subsection{Sequential Mode Decoding}
\label{sec:decoder}
To facilitate reasoning about multimodality in the absence of multimodal ground truth, our framework requires the decoder to conduct chain-based factorization on the joint mode embeddings, which is demonstrated as follows:
\begin{equation}
    \bm{m}_{k} = \decoder\left(\Psi,\, \{\bm{m}_1,\ldots,\bm{m}_{k-1}\}\right),\, k \in \{1,\ldots,K\}\,.
\label{eq:seq}
\end{equation}
With such a factorization that converts the unordered set of modes into a sequence, the correlation between modes can be naturally strengthened, as the mode to be decoded depends on the ones that appear previously. Further equipping this framework with appropriate model implementation and training scheme has the potential to offer better mode coverage and scoring without severe sacrifice in trajectory accuracy, which we introduce in the following sections.

\subsection{Single-Layer Mode Sequence}
\label{sec:single}

This section illustrates the detailed structure of a single ModeSeq layer, which consists of a Memory Transformer module and a Context Transformer module. Stacking multiple ModeSeq layers can further improve the performance via iterative refinement, which will be discussed in \cref{sec:multi}.

\noindent \textbf{Routine.} We introduce the decoding procedure of the $\ell$-th ModeSeq layer, with the right-hand side of \cref{fig:modeseq} depicting the first layer (\ie, $\ell=1$) for a single agent. Note that the computation graph of the layer can be executed for multiple agents in parallel. The layer is designed to recurrently output a sequence of mode embeddings $[\bm{m}_1^{(\ell)},\ldots,\bm{m}_K^{(\ell)}]$, where we slightly complicate the notation of modes with a superscript that identifies the layer index. The input of the layer includes the scene embedding $\Psi$ yielded by the encoder and the mode embeddings produced by the $(\ell-1)$-th decoding layer, \ie, $[\bm{m}_1^{(\ell-1)},\ldots,\bm{m}_K^{(\ell-1)}]$. Since the latter input does not apply to the first layer, we introduce an embedding $\bm{e} \in \mathbb{R}^D$ to serve as the output of the ``$0$-th layer'', which is randomly initialized at the beginning of training. The same learnable $\bm{e}$ is shared across the $K$ input embeddings of the first layer:
\vspace{-0.2cm}
\begin{equation}
    \bm{m}_k^{(0)} = \bm{e},\, k \in \{1,\ldots,K\}\,.
    \vspace{-0.2cm}
\end{equation}
Before starting the decoding, we create an empty sequence $\Omega_0^{(\ell)} = [\,]$ with the subscript and superscript indicating the $0$-th decoding step and the $\ell$-th layer, respectively. This sequence will be used to keep track of the mode embeddings produced at various steps. At the $k$-th decoding step, we employ a Memory Transformer module and a Context Transformer module to update $\bm{m}_{k}^{(\ell-1)}$ to become $\bm{m}_{k}^{(\ell)}$, leveraging the information in the memory bank $\Omega_{k-1}^{(\ell)}$ and the scene embedding $\Psi$. The output mode embedding $\bm{m}_{k}^{(\ell)}$ is then pushed to the end of the sequence $\Omega_{k-1}^{(\ell)}$ to obtain $\Omega_{k}^{(\ell)} = [\bm{m}_1^{(\ell)},\ldots,\bm{m}_{k}^{(\ell)}]$, which will serve as the input at the $(k+1)$-th decoding step. After going through $K$ decoding steps, we use a prediction head to transform each of the mode embeddings stored in $\Omega_K^{(\ell)}$ into a specific trajectory and a corresponding confidence score via MLPs. The following paragraphs detail the modules constituting a ModeSeq layer and discuss the differences between our approach and other alternatives.

\noindent \textbf{Memory Transformer.} The Memory Transformer takes charge of modeling the sequential dependencies of trajectory modes. At the $k$-th decoding step of the $\ell$-th layer, this module takes as input $\Omega_{k-1}^{(\ell)}$ and $\bm{m}_k^{(\ell-1)}$, the memory bank for the current layer and the $k$-th mode embedding derived from the last layer. Since we desire the generation of the $k$-th mode embedding $\bm{m}_k^{(\ell)}$ to be aware of the existence of the preceding modes, we treat $\bm{m}_k^{(\ell-1)}$ as the query of the Transformer module, which retrieves the memory bank in a cross-attention manner:
\vspace{-0.1cm}
\begin{equation}
    \hat{\bm{m}}_k^{(\ell)} = \memory\left(\text{query}{=}\bm{m}_k^{(\ell-1)},\,\text{key}/\text{value}{=}\Omega_{k-1}^{(\ell)}\right)\,.
    \vspace{-0.1cm}
\end{equation}
In this way, the information in $\Omega_{k-1}^{(\ell)}$ is assimilated into $\bm{m}_k^{(\ell-1)}$ to produce $\hat{\bm{m}}_k^{(\ell)}$, which is a query feature conditioned on the modes up to the $(k-1)$-th decoding step.

\noindent \textbf{Context Transformer.} To derive scene-compliant modes, we must provide the query feature with the specific scene context. To this end, we use the Context Transformer module to refine the conditional query $\hat{\bm{m}}_k^{(\ell)}$ with the scene embeddings output by the encoder. Specifically, the $k$-th mode embedding $\bm{m}_k^{(\ell)}$ is computed by enriching $\hat{\bm{m}}_k^{(\ell)}$ with $\Psi$ using cross-attention:
\vspace{-0.2cm}
\begin{equation}
    \bm{m}_k^{(\ell)} = \context\left(\text{query}=\hat{\bm{m}}_k^{(\ell)},\,\text{key}/\text{value}=\Psi\right)\,.
    \vspace{-0.1cm}
\end{equation}
Considering the high complexity of performing global attention, we decompose the Context Transformer into three separate modules in practice, including mode-time cross-attention, mode-map cross-attention, and mode-agent cross-attention, each of which takes as input only a subset of the embeddings contained in $\Psi$. First, the mode-time cross-attention fuses the query feature with the historical encoding belonging to the agent of interest, enabling the query to adapt to the specific agent. Second, we aggregate the map information surrounding the agent of interest into the query feature leveraging the mode-map cross-attention, which contributes to the map compliance of the forecasting results. Finally, utilizing the mode-agent cross-attention module to fuse the neighboring agents' embeddings at the current time step promotes the model's social awareness. After going through these three modules, the conditional query $\hat{\bm{m}}_k^{(\ell)}$ eventually becomes $\bm{m}_k^{(\ell)}$, which is now context-aware.

\noindent \textbf{Prediction Head.} Given the conditional, context-aware mode embedding $\bm{m}_k^{(\ell)}$, we use an MLP head to output the $k$-th trajectory $\hat{\bm{y}}_k^{(\ell)}$ and another to estimate the corresponding confidence score $\hat{\bm{\phi}}_k^{(\ell)}$:
\vspace{-0.2cm}
\begin{equation}
\begin{dcases}
    \hat{\bm{y}}_k^{(\ell)} = \mlp\left(\bm{m}_k^{(\ell)}\right)\,, \\
    \hat{\bm{\phi}}_k^{(\ell)} = \mlp\left(\bm{m}_k^{(\ell)}\right)\,. \\
\end{dcases}
\vspace{-0.2cm}
\end{equation}

\noindent \textbf{Comparison with DETR-Like Decoders.} In contrast to motion decoders~\cite{varadarajan2022multipath++, nayakanti2022wayformer, zhou2023query, shi2022mtr} inspired by DETR~\cite{carion2020end}, where the relationships between modes are completely neglected~\cite{varadarajan2022multipath++, nayakanti2022wayformer} or weakly modeled by mode-mode self-attention~\cite{zhou2023query, shi2022mtr}, modeling modes as a sequence strengthens mode-wise relational reasoning thanks to the conditional dependence in generating multimodal embeddings, which is beneficial to eliminating duplicated trajectories. Furthermore, DETR-like approaches can only decode a fixed number of modes, as the number of learnable/static anchors cannot be changed once specified at the start of training. By contrast, our ModeSeq framework completely achieves parameter sharing across decoding steps and supports outputting more/fewer modes at test time, which can be simply achieved by changing the number of decoding steps. This characteristic can be helpful since the degree of uncertainty varies by scenario.

\noindent \textbf{Comparison with Typical Recurrent Networks.} The ModeSeq layer can be viewed as a sort of recurrent network~\cite{hochreiter1997long, cho2014learning} due to its parameter sharing across decoding steps, though its core components are modernized with Transformers. While typical recurrent networks compress the memory into a single hidden state, which is often lossy, the Memory Transformer in a ModeSeq layer allows for direct access to all prior mode embeddings, naturally scaling the capacity of the memory as the number of modes grows.

\subsection{Multi-Layer Mode Sequences}
\label{sec:multi}

Single-layer mode sequences may have limited capability of learning high-quality mode representations. In particular, if the layer happens to produce unrealistic or less likely modes at the first few decoding steps, the learning of the later modes may be unexpectedly disturbed. Inspired by DETR~\cite{carion2020end}, we develop an iterative refinement framework by stacking multiple ModeSeq layers and applying training losses to the output of each layer. As shown in the left part of \cref{fig:modeseq}, all layers except for the first one take as input the mode embeddings output from the last round of decoding, refining the features with the scene context. Crucially, we introduce the operation of mode rearrangement in between layers, which corrects the order of the embeddings in the mode sequence to encourage decoding trajectory modes with monotonically decreasing confidence scores.

\noindent \textbf{Mode Rearrangement.} Before transitioning from the $\ell$-th to the $(\ell+1)$-th ModeSeq layer, we sort the mode embeddings stored in the memory bank $\Omega_K^{(\ell)}$ according to the descending order of the confidence scores predicted from them. The sorted mode embeddings will then be sequentially input to the $(\ell+1)$-th ModeSeq layer for recurrent decoding. Through iterative refinement with mode rearrangement, the trajectories and the order of modes become more scene-compliant and more monotonous, respectively.

\subsection{Early-Match-Take-All Training}
\label{sec:train}

The WTA training strategy~\cite{lee2016stochastic} is blamed for producing overlapped trajectories and indistinguishable confidence scores~\cite{lin2024eda, makansi2019overcoming, rupprecht2017learning, thiede2019analyzing}. Fortunately, our approach has the opportunity to mitigate this issue by opting for a more appropriate training method under the paradigm of sequential mode modeling. In this section, we propose the EMTA loss, which leverages the order of modes to define the positive and negative samples toward better mode coverage and confidence scoring without significantly sacrificing trajectory accuracy.

\begin{table*}[t]
\footnotesize
\centering
\caption{Quantitative results on the 2024 Waymo Open Dataset Motion Prediction Benchmark.}
\vspace{-0.3cm}
\setlength{\tabcolsep}{4.1mm}
\begin{tabular}{@{}c|l|cc|ccccc@{}}
\toprule
Dataset & Method & Ensemble & Lidar & Soft mAP${}_6$ $\uparrow$ & mAP${}_6$ $\uparrow$ & MR${}_6$ $\downarrow$ & minADE${}_6$ $\downarrow$ & minFDE${}_6$ $\downarrow$ \\
\midrule
\multirow{6}*{Val} & MTR v3~\cite{mtrv3} & $\times$ & \checkmark & - & 0.4593 & 0.1175 & 0.5791 & 1.1809 \\
& MTR++~\cite{shi2023mtr++} & $\times$ & $\times$ & - & 0.4382 & 0.1337 & 0.6031 & 1.2135 \\
& QCNet~\cite{zhou2023query} & $\times$ & $\times$ & 0.4508 & 0.4452 & 0.1254 & 0.5122 & 1.0225 \\
& ModeSeq (Ours) & $\times$ & $\times$ & 0.4562 & 0.4507 & 0.1206 & 0.5237 & 1.0681 \\
\midrule
\multirow{4}*{Test} & MTR v3~\cite{mtrv3} & \checkmark & \checkmark & 0.4967 & 0.4859 & 0.1098 & 0.5554 & 1.1062 \\
& RMP\_Ensemble~\cite{sun2024rmp} & \checkmark & $\times$ & 0.4726 & 0.4553 & 0.1113 & 0.5596 & 1.1272 \\ 
& ModeSeq (Ours) & \checkmark & $\times$ & 0.4737 & 0.4665 & 0.1204 & 0.5680 & 1.1766 \\ 
& ModeSeq (Ours) & $\times$ & $\times$ & 0.4487 & 0.4450 & 0.1244 & 0.5304 & 1.0836 \\ 
\bottomrule
\end{tabular}
\label{tab:waymo}
\vspace{-0.6cm}
\end{table*}

\begin{table}[t]
\footnotesize
\centering
\caption{Quantitative results on the 2024 Argoverse 2 Single-Agent Motion Forecasting Benchmark.}
\vspace{-0.3cm}
\setlength{\tabcolsep}{0.2mm}
\begin{tabular}{@{}l|c|ccccccc@{}}
\toprule
Method & Ensemble & b-minFDE${}_6$ $\downarrow$ & MR${}_6$ $\downarrow$ & minADE${}_6$ $\downarrow$ & minFDE${}_6$ $\downarrow$ \\
\midrule
MTR~\cite{shi2022mtr} & \checkmark & 1.98 & 0.15 & 0.73 & 1.44 \\
MTR++~\cite{shi2023mtr++} & \checkmark & 1.88 & 0.14 & 0.71 & 1.37 \\
QCNet~\cite{zhou2023query} & $\times$ & 1.91 & 0.16 & 0.65 & 1.29 \\
ModeSeq (Ours) & $\times$ & 1.87 & 0.14 & 0.63 & 1.26 \\
\bottomrule
\end{tabular}
\label{tab:argoverse}
\vspace{-0.6cm}
\end{table}

Typical WTA loss optimizes only the trajectory with the minimum displacement error with respect to the ground truth. In comparison, our EMTA loss optimizes the \textit{matched} trajectory decoded at the \textit{earliest} recurrent step. For example, if both the second and the third trajectories match the ground truth, only the second one will be optimized, regardless of which one has the minimum error. To this end, we search over the $K$ predictions to acquire the collection of mode indices associated with matched trajectories:
\vspace{-0.2cm}
\begin{equation}
    G^{(\ell)} = \left\{k \mid k \in\left\{1,\cdots,K\right\} \wedge \mathds{1}\left\{\match\left(\hat{\bm{y}}_k^{(\ell)},\, \bm{y}\right)\right\} \right\}\,,
    \vspace{-0.1cm}
\end{equation}
where $G^{(\ell)}$ denotes the set of qualified mode indices in the $\ell$-th ModeSeq layer, $\mathds{1}\{\cdot\}$ represents the indicator function, and $\match(\cdot,\,\cdot)$ defines the criterion for a match given the ground-truth trajectory $\bm{y}$. The implementation of $\match(\cdot,\,\cdot)$ can be flexible, depending on the trajectory accuracy demanded by practitioners. For instance, on the Waymo Open Motion Dataset~\cite{ettinger2021large}, we decide whether a predicted trajectory is a match based on the velocity-aware distance thresholds defined in the Miss Rate metric of the benchmark; while on the Argoverse 2 Motion Forecasting Dataset~\cite{Argoverse2}, a matched trajectory is expected to have less than $2$-meter final displacement error. Given $G^{(\ell)}$, we determine the unique positive sample's index $\hat{k}^{(\ell)}$ as follows, with all the remaining modes treated as negative samples:
\vspace{-0.2cm}
\begin{equation}
    \hat{k}^{(\ell)} = 
    \begin{dcases}
    \vspace{-0.2cm}
        \min_{k \in G^{(\ell)}} k & \text{if } |G^{(\ell)}| > 0 \,;\\
        \operatorname*{argmin}_{k} \distance\left(\hat{\bm{y}}_k^{(\ell)},\, \bm{y}\right) & \text{otherwise} \,, \\
    \end{dcases}
    \vspace{-0.2cm}
\end{equation}
where $|\cdot|$ denotes the cardinality of a set, and $\distance(\cdot,\,\cdot)$ measures the average displacement error between trajectories. This strategy for label assignment encourages the model to decode matched trajectories as early as possible by treating the earliest instead of the best matches as positive samples. Meanwhile, it drives the later matches, if any, away from the ground truth by assigning negative labels to them. On the other hand, if none of the predictions match, which commonly happens at the early stage of training, we will fall back to the regular WTA scheme to ease the difficulty in optimization. Following label assignment, we use the Laplace negative log-likelihood~\cite{zhou2022hivt, zhou2023query} as the regression loss, optimizing the trajectories of the positive samples. Besides, we use the Binary Focal Loss~\cite{lin2017focal} to optimize the confidence scores according to the labels assigned. We also try a variant of confidence loss, where we introduce the definition of ignored samples to mask the confidence loss of the modes decoded earlier than the positive samples, which is shown to be effective in the absence of mode rearrangement.

\section{Experiments}

\begin{table*}[t]
\footnotesize
\centering
\caption{Effects of sequential mode modeling and Early-Match-Take-All training on the validation set of the WOMD.}
\vspace{-0.3cm}
\setlength{\tabcolsep}{3.6mm}
\begin{tabular}{@{}c|c|c|ccccc@{}}\toprule
Decoder & Training Strategy & Ignored Samples & Soft mAP${}_6$ $\uparrow$ & mAP${}_6$ $\uparrow$ & MR${}_6$ $\downarrow$ & minADE${}_6$ $\downarrow$ & minFDE${}_6$ $\downarrow$ \\
\midrule
\multirow{2}*{DETR w/ Refinement} & \multirow{2}*{WTA} & None & 0.4096 & 0.4050 & 0.1536 & 0.5660 & 1.1716 \\
& & Other Matches & \textbf{0.4150} & \textbf{0.4103} & \textbf{0.1502} & \textbf{0.5619} & \textbf{1.1621} \\
\midrule
\multirow{5}*{ModeSeq (Ours)} & \multirow{2}*{WTA} & None & 0.4138 & 0.4093 & \textbf{0.1502} & 0.5563 & \textbf{1.1498} \\
& & Other Matches & \textbf{0.4207} & \textbf{0.4161} & 0.1503 & \textbf{0.5556} & 1.1501 \\
\cmidrule{2-8}
& \multirow{2}*{EMTA} & None & \textbf{0.4231} & \textbf{0.4196} & \textbf{0.1457} & \textbf{0.5700} & \textbf{1.1851} \\
& & Other Matches & 0.4098 & 0.4060 & 0.1496 & 0.5817 & 1.2207 \\
\bottomrule
\end{tabular}
\vspace{-0.3cm}
\label{tab:emta}
\end{table*}

\begin{table*}[t]
\footnotesize
\centering
\caption{Effects of mode rearrangement on the validation set of the WOMD.}
\vspace{-0.3cm}
\setlength{\tabcolsep}{5.8mm}
\begin{tabular}{@{}c|c|ccccc@{}}
\toprule
Mode Rearrangement & Ignored Samples & Soft mAP${}_6$ $\uparrow$ & mAP${}_6$ $\uparrow$ & MR${}_6$ $\downarrow$ & minADE${}_6$ $\downarrow$ & minFDE${}_6$ $\downarrow$ \\
\midrule
\multirow{2}*{$\times$} & None & 0.4112 & 0.4077 & 0.1548 & 0.5884 & 1.2389 \\
& Early Mismatches & \textbf{0.4141} & \textbf{0.4109} & \textbf{0.1489} & \textbf{0.5749} & \textbf{1.2066} \\
\midrule
\multirow{2}*{\checkmark} & None & \textbf{0.4231} & \textbf{0.4196} & \textbf{0.1457} & \textbf{0.5700} & \textbf{1.1851} \\
& Early Mismatches & 0.4161 & 0.4129 & 0.1461 & 0.5751 & 1.2041 \\
\bottomrule
\end{tabular}
\label{tab:rearrangement}
\vspace{-0.6cm}
\end{table*}

\subsection{Experimental Setup}
\noindent \textbf{Datasets.} We conduct experiments on the Waymo Open Motion Dataset (WOMD)~\cite{ettinger2021large} and the Argoverse 2 Motion Forecasting Dataset~\cite{Argoverse2}. The WOMD contains $486995$/$44097$/$44920$ training/validation/testing samples, where the history of $1.1$ seconds is provided as the context and the $8$-second future trajectories of up to $8$ agents are required to predict. The Argoverse 2 dataset comprises $199908$/$24988$/$24984$ samples with $5$-second observation windows and $6$-second prediction horizons for training/validation/testing.

\noindent \textbf{Metrics.} Following the standard of the benchmarks~\cite{ettinger2021large, Argoverse2}, we constrain models to output at most $K=6$ trajectories. We use Miss Rate (MR${}_K$) to measure mode coverage, which counts the fraction of cases in which the model fails to produce any trajectories that match the ground truth within the required thresholds. Built upon the definition of a match, mAP${}_K$ and Soft mAP${}_K$ assess the quality of the confidence scores by computing the P/R curves and averaging the precision values over various confidence thresholds. If multiple predictions match the ground truth, only the one with the highest confidence is deemed true-positive, so gathering multiple predictions around the same region with high confidence may introduce more false positives and lower precision. Moreover, if the confidence of a matched mode is below the threshold, the recall would be harmed. Since mAP${}_K$ and Soft mAP${}_K$ penalize high-confidence unmatched and low-confidence matched modes, a model must predict diverse trajectories with calibrated confidences to score high on these metrics. To further evaluate trajectory quality, we use minimum Average Displacement Error (minADE${}_K$) and minimum Final Displacement Error (minFDE${}_K$) as indicators, which calculate the distance between the ground truth and the best-predicted trajectories as an average over the whole horizon and at the final time step, respectively. Finally, the b-minFDE${}_K$ concerns the joint performance of trajectories and confidences by summing the minFDE${}_K$ and the Brier scores of the best modes.

\noindent \textbf{Implementation Details.} We develop models with a hidden size of $128$. The decoder stacks $6$ layers for iterative refinement, with each layer executing $6$ steps to obtain exactly $6$ modes as required by the benchmarks~\cite{ettinger2021large, Argoverse2}. On the WOMD~\cite{ettinger2021large}, we use the AdamW optimizer~\cite{loshchilov2017decoupled} to train models for $30$ epochs on the training set with a batch size of $32$, a weight decay rate of 0.1, and a dropout rate of $0.1$. On Argoverse 2~\cite{Argoverse2}, we use a similar training configuration except that the number of epochs is extended to $64$. The initial learning rate is set to $5 \times 10^{-4}$, which is decayed to $0$ at the end of training following the cosine annealing schedule~\cite{loshchilov2016sgdr}. Unless specified, the ablation studies are based on experiments on the WOMD with $20\%$ of the training data.

\subsection{Comparison with State of the Art}

We compare our approach with QCNet~\cite{zhou2023query} and the MTR series~\cite{shi2022mtr, shi2023mtr++, mtrv3}, which are currently the most effective sparse and dense multimodal prediction solutions across the WOMD and the Argoverse 2 dataset. As demonstrated in \cref{tab:waymo}, ModeSeq achieves the best scoring performance among the Lidar-free methods on the validation split of the WOMD, though it lags behind the mAP${}_6$ performance of MTR v3~\cite{mtrv3}, a model that augments the input information with raw sensor data. As a sparse mode predictor, ModeSeq undoubtedly outperforms MTR++~\cite{shi2023mtr++} in terms of MR${}_6$, minADE${}_6$, and minFDE${}_6$ by a large margin. Compared with QCNet~\cite{zhou2023query}, ModeSeq attains better Soft mAP${}_6$, mAP${}_6$, and MR${}_6$ at the cost of slight degradation on minADE${}_6$ and minFDE${}_6$, confirming that our approach can improve the mode coverage and confidence scoring of sparse predictors without significant sacrifice in trajectory accuracy. In the 2024 Waymo Open Motion Prediction Challenge, the ensemble version of ModeSeq ranked first among Lidar-free approaches on the test set of the WOMD. Our approach also exhibits promising performance on the Argoverse 2 dataset, where our ensemble-free model surpasses QCNet and the MTR series on all critical metrics as shown in \cref{tab:argoverse}.

\subsection{Ablation Study}

\begin{figure}
  \centering
  \includegraphics[width=1.0\linewidth]{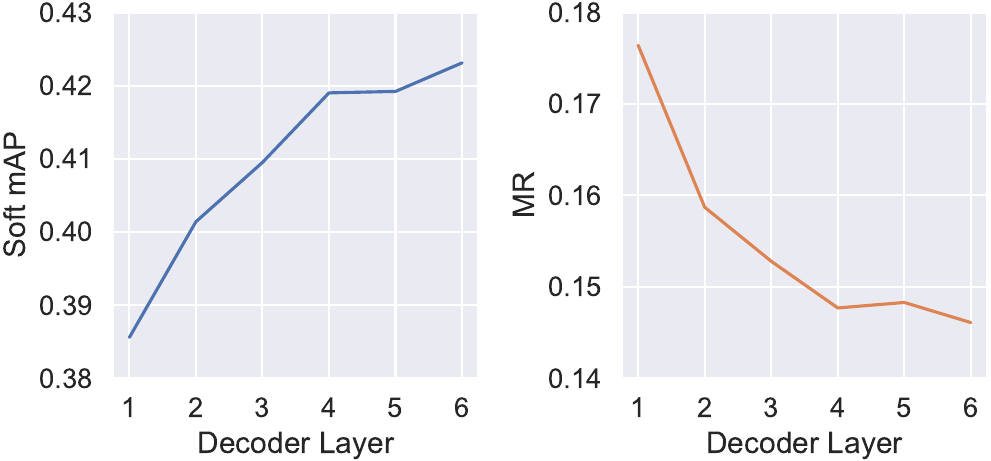}
  \vspace{-0.7cm}
  \caption{The performance after each decoding layer on the validation set of the WOMD.}
  \label{fig:refinement}
  \vspace{-0.7cm}
\end{figure}

\begin{figure*}
  \centering
  \begin{subfigure}{0.33\linewidth}
    \includegraphics[width=1.0\textwidth,height=0.75\textwidth]{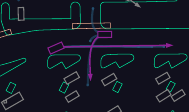}
    \caption{\#Mode@Training=3, \#Mode@Inference=3}
    \label{fig:train_3_infer_3}
  \end{subfigure}
  \begin{subfigure}{0.33\linewidth}
    \includegraphics[width=1.0\textwidth,height=0.75\textwidth]{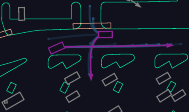}
    \caption{\#Mode@Training=6, \#Mode@Inference=6}
    \label{fig:train_6_infer_6}
  \end{subfigure}
  \begin{subfigure}{0.33\linewidth}
    \includegraphics[width=1.0\textwidth,height=0.75\textwidth]{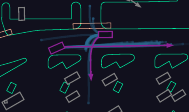}
    \caption{\#Mode@Training=6, \#Mode@Inference=24}
    \label{fig:train_6_infer_24}
  \end{subfigure}
  \vspace{-0.7cm}
  \caption{Visualization on the WOMD. The agents in purple are predicted with blue trajectories, with the opacity indicating confidence.}
  \vspace{-0.6cm}
  \label{fig:qualitative}
\end{figure*}

\noindent \textbf{Effects of Sequential Mode Modeling.} In \cref{tab:emta}, we examine the effectiveness of sequential mode modeling by comparing ModeSeq with the sparse DETR-like decoder enhanced with iterative refinement~\cite{carion2020end}, both employing the same QCNet encoder~\cite{zhou2023query} for fair comparisons. The results demonstrate that ModeSeq outperforms the baseline on all metrics when using the same training strategy. Interestingly, ignoring the confidence loss of the suboptimal modes that match the ground truth can improve the performance of both methods under the WTA training. The reason behind this is that treating the other matched modes as negative samples will confuse the optimization process, given that the best and the other matches usually have similar mode representations while they are assigned as opposite samples.

\noindent \textbf{Effects of EMTA Training.} We also investigate the role of EMTA training in \cref{tab:emta}. After replacing the WTA loss with our EMTA scheme, the results on Soft mAP${}_6$, mAP${}_6$, and MR${}_6$ are considerably improved, which demonstrates the benefits of EMTA training in terms of mode coverage and confidence scoring. On the other hand, the performance on minADE${}_6$ and minFDE${}_6$ slightly deteriorates since the EMTA loss has relaxed the requirement for trajectory accuracy, but the degree of deterioration is less than $0.02$ meters in minADE${}_6$ and $0.04$ meters in minFDE${}_6$, leading to more balanced performance taken overall. Contrary to the conclusion drawn from the WTA baselines, treating other matches as ignored samples is detrimental under the EMTA strategy. This is because the joint effects of sequential mode decoding and EMTA training have broken the symmetry of mode modeling and label assignment, allowing us to assign the other matches as negative samples to drive them away from the ground truth for covering other likely modes.

\noindent \textbf{Effects of Iterative Refinement.} To understand the effects of iterative refinement under our framework, we take the output from different decoding layers for evaluation. As shown in \cref{fig:refinement}, the performance on Soft mAP${}_6$ and MR${}_6$ is generally improved as the depth increases, totaling a substantial enhancement between the first and the last layer. One of the reasons why iterative refinement works well can be attributed to the operation of mode rearrangement in between layers, which we explain in the following.

\noindent \textbf{Effects of Mode Rearrangement.} Mode rearrangement coordinates with EMTA training to facilitate decoding matched trajectories with high confidence as early as possible. Comparing the first and third rows of \cref{tab:rearrangement}, we can see that reordering the mode embeddings before further refinement can remarkably promote the forecasting capability. To gain deeper insights into the results, we develop a variant of label assignment, where the modes decoded earlier than the first match are deemed ignored samples. Again, incorporating mode rearrangement is beneficial under this setting according to the second and fourth rows of \cref{tab:rearrangement}. Interestingly, we found this strategy of label assignment to outperform the default one in the absence of mode rearrangement (see the first half of \cref{tab:rearrangement}). This phenomenon can be explained by the fact that bad modes may appear in the first few decoding steps of the shallow layers, which can negatively impact the learning of the subsequent modes. To implicitly guide the model to output more confident modes first, we can provide monotonically decreasing confidence labels by ignoring the confidence loss of the early mismatches. Playing a similar but stronger role, mode rearrangement explicitly prioritizes the refinement of the more probable trajectories in the next layer by manually placing the less confident modes at the end of the sequence.

\noindent \textbf{Capability of Representative Mode Learning.} We demonstrate ModeSeq's ability to produce representative modes in \cref{tab:representative}. While training models to decode merely $3$ modes necessarily leads to worse performance, the $3$-mode variant of ModeSeq achieves the same level of performance on Soft mAP${}_6$ and mAP${}_6$ compared with the $6$-mode model. By comparison, QCNet~\cite{zhou2023query} fails to achieve comparable results if only using $3$ mode queries during training.

\begin{figure}
  \centering
  \includegraphics[width=1.0\linewidth]{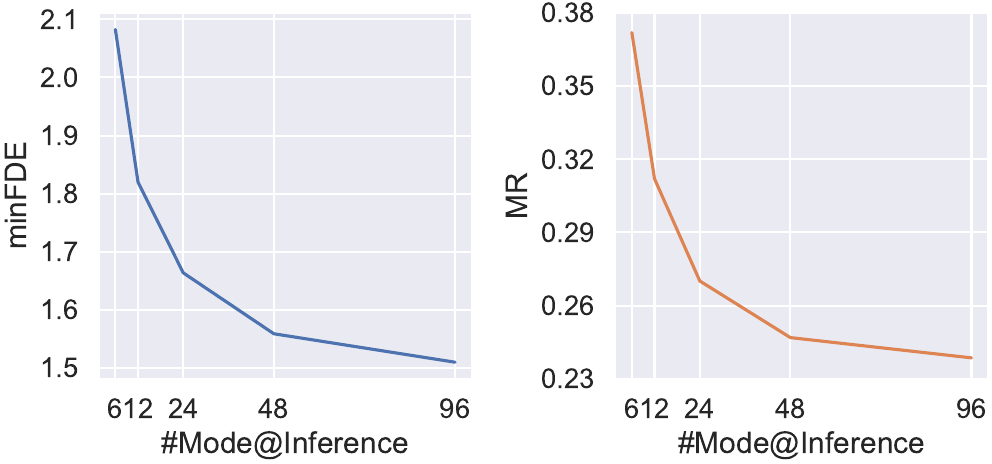}
  \vspace{-0.8cm}
  \caption{The results of generating more than $6$ modes on the validation set of the WOMD.}
  \label{fig:extrapolation}
  \vspace{-0.6cm}
\end{figure}

\noindent \textbf{Capability of Mode Extrapolation.} We ask the model trained by generating $6$ modes to execute more decoding steps at test time. As depicted in \cref{fig:extrapolation}, ModeSeq achieves lower prediction error with the increase of the decoded modes, emerging with the capability of mode extrapolation thanks to sequential modeling. This characteristic enables handling various degrees of uncertainty across scenarios.

\noindent \textbf{Inference Cost.} As shown in \cref{tab:representative}, the inference latency of ModeSeq is about twice as high as that of QCNet~\cite{zhou2023query} if predicting $6$ modes. However, in many industrial autonomous driving solutions, the number of modes used by downstream planning is refrained from exceeding $3$. \Cref{tab:representative} shows that the latency gap between the $3$-mode models is much smaller, while the mAP${}_6$ of $3$-mode ModeSeq is even better than that of $6$-mode QCNet. Given our approach's capability of producing representative trajectories with fewer modes, we believe sequential mode modeling has the potential to be deployed on board. 


\subsection{Qualitative Results}

The qualitative results produced by ModeSeq are presented in \cref{fig:qualitative}. \Cref{fig:train_3_infer_3} demonstrates that our model can generate representative trajectories when being trained to decode only $3$ modes. Comparing \cref{fig:train_6_infer_24} with \cref{fig:train_6_infer_6}, we can see that the $6$-mode model successfully extrapolates diverse yet realistic modes when executing $24$ decoding steps during inference, showcasing the extrapolation ability of ModeSeq.



\begin{table}[t]
\footnotesize
\centering
\caption{Capability of generating representative modes with precise confidence scores. Models are trained on $100\%$ training data and evaluated on the validation split of the WOMD.}
\vspace{-0.4cm}
\setlength{\tabcolsep}{0.9mm}
\begin{tabular}{@{}c|c|ccc|c@{}}
\toprule
\#Mode & Model & Soft mAP${}_6$ $\uparrow$ & mAP${}_6$ $\uparrow$ & MR${}_6$ $\downarrow$ & Latency (ms) \\
\midrule
\multirow{2}*{3} & QCNet~\cite{zhou2023query} & 0.4214 & 0.4163 & 0.2007 & 63$\pm$11 \\
& ModeSeq (Ours) & 0.4509 & 0.4479 & 0.1967 & 86$\pm$9 \\
\hline
\multirow{2}*{6} & QCNet~\cite{zhou2023query} & 0.4508 & 0.4452 & 0.1254 & 69$\pm$16 \\
& ModeSeq (Ours) & 0.4562 & 0.4507 & 0.1206 & 143$\pm$10 \\
\bottomrule
\end{tabular}
\vspace{-0.4cm}
\label{tab:representative}
\end{table}

\section{Conclusion}
This paper introduces ModeSeq, a modeling framework that achieves sparse multimodal motion prediction via sequential mode modeling. The framework comprises a mechanism of sequential multimodal decoding, an architecture of iterative refinement with mode rearrangement, and a training strategy of Early-Match-Take-All label assignment. As an alternative to the unordered multimodal decoding and the winner-take-all training strategy, ModeSeq achieves state-of-the-art results on motion prediction benchmarks and exhibits the characteristic of mode extrapolation, creating a new path to solving multimodal problems.


\section*{Acknowledgement} This project is supported by a grant from Hong Kong Research Grant Council under CRF C1042-23G.
{
    \small
    \bibliographystyle{ieeenat_fullname}
    \bibliography{main}
}

\clearpage
\setcounter{page}{1}
\maketitlesupplementary


\section{Definition of a Match}

The Argoverse 2 Motion Forecasting Benchmark~\cite{Argoverse2} desires the predictions' displacement error at the $60$-th time step to be less than $2$ meters. By linearly scaling the $2$-meter threshold across time steps, we obtain a distance threshold $\Gamma(t)$ for each time step $t$:
\begin{equation}
    \Gamma\left(t\right) = \frac{t}{30}\,.
\end{equation}
Our EMTA training loss requires a matched trajectory to fall within the given threshold of the ground truth at every future time step.

On the Waymo Open Motion Dataset (WOMD)~\cite{ettinger2021large}, the thresholds are divided into lateral and longitudinal ones, which are adaptive to the current velocity of the agent of interest. To begin with, the benchmark defines a scaling factor with respect to the velocity $\bm{v}$:
\begin{equation}
    \operatorname*{Scale}\left(\bm{v}\right) =
    \begin{dcases}
        0.5 & \text{if } \bm{v} < 1.4 \,;\\
        0.5 + 0.5 \frac{\bm{v} - 1.4}{11 - 1.4} & \text{if } 1.4 \leq \bm{v} < 11 \,;\\
        1 & \text{otherwise} \,. \\
    \end{dcases}
\end{equation}
Utilizing the scaling factor, we define the lateral threshold $\Gamma_{\text{lat}}(\bm{v},\,t)$ as
\begin{equation}
    \Gamma_{\text{lat}}\left(\bm{v},\,t\right) = \operatorname*{Scale}\left(\bm{v}\right) \times
    \begin{dcases}
        \frac{t}{30} & \text{if } 1 \leq t \leq 30 \,;\\
        0.04 t - 0.2 & \text{otherwise} \,. \\
    \end{dcases}
    \label{eq:lat}
\end{equation}
Similarly, we set the longitudinal threshold $\Gamma_{\text{lon}}(\bm{v},\,t)$ to be twice as large as the lateral one:
\begin{equation}
    \Gamma_{\text{lon}}\left(\bm{v},\,t\right) = \operatorname*{Scale}\left(\bm{v}\right) \times
    \begin{dcases}
        \frac{t}{15} & \text{if } 1 \leq t \leq 30 \,;\\
        0.08 t - 0.4 & \text{otherwise} \,. \\
    \end{dcases}
    \label{eq:lon}
\end{equation}
Regarding the experiments on the WOMD, we demand a matched trajectory to have errors below both the lateral and longitudinal thresholds at every future time step.

\section{Ensemble Method on the WOMD}

Inspired by Weighted Boxes Fusion (WBF)~\cite{solovyev2021weighted}, we propose Weighted Trajectory Fusion to aggregate multimodal trajectories produced by multiple models. Our ensemble method is almost the same as WBF, except we are fusing trajectories according to distance thresholds rather than bounding boxes according to IOU thresholds. Our ensemble method can improve mAP${}_6$/Soft mAP${}_6$/MR${}_6$ by sacrificing minADE${}_6$/minFDE${}_6$, which indicates that the performance on various metrics sometimes disagrees. The critical hyperparameters in Weighted Trajectory Fusion are the distance thresholds used for trajectory clustering. We choose the velocity-aware thresholds defined in \cref{eq:lat} and \cref{eq:lon} as the base thresholds. On top of this, we multiply the base thresholds with the scaling factors of $1.5$, $1.4$, and $1.4$ for vehicles, pedestrians, and cyclists, respectively.

\section{Versatility}

\begin{table}[t]
\footnotesize
\centering
\caption{Applying ModeSeq to other scene encoders on the WOMD. Models are trained on $20\%$ training data and evaluated on the validation set.}
\setlength{\tabcolsep}{1.5mm}
\begin{tabular}{@{}l|ccccc@{}}
\toprule
Model & Soft mAP${}_6$ $\uparrow$ & mAP${}_6$ $\uparrow$ & MR${}_6$ $\downarrow$ \\
\midrule
Scene Transformer~\cite{ngiam2021scene} & 0.2441 & 0.2416 & 0.2977 \\
Scene Transformer~\cite{ngiam2021scene} + ModeSeq & \textbf{0.4130} & \textbf{0.4095} & \textbf{0.1513} \\
\hline
HiVT~\cite{zhou2022hivt} & 0.3576 & 0.3536 & 0.1911 \\
HiVT~\cite{zhou2022hivt} + ModeSeq & \textbf{0.4121} & \textbf{0.4086} & \textbf{0.1485} \\
\bottomrule
\end{tabular}
\label{tab:other_encoder}
\end{table}

Our ModeSeq framework can be seamlessly integrated with other scene encoders. As shown in~\cref{tab:other_encoder}, our approach significantly enhances Scene Transformer~\cite{ngiam2021scene} and HiVT~\cite{zhou2022hivt}, two representative methods adopting scene-centric and agent-centric encoders, respectively. Currently, our EMTA training strategy can only be applied to mode embeddings with order since the definition of ``early match'' relies on order.

\section{Parameter Efficiency}

The ModeSeq decoder comprises $7.5$M parameters, totaling $10.9$M parameters for the overall model when combined with the QCNet encoder~\cite{zhou2023query}, which is far more parameter-efficient than other state-of-the-art on the WOMD (\eg, MTR~\cite{shi2022mtr} with $65.8$M parameters and MTR++~\cite{shi2023mtr++} with $86.6$M parameters).

\section{More Qualitative Results}

\begin{figure*}[t]
  \centering
  \begin{subfigure}{0.33\linewidth}
    \includegraphics[width=1.0\textwidth,height=0.75\textwidth]{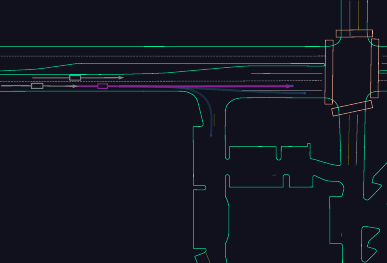}
    \includegraphics[width=1.0\textwidth,height=0.75\textwidth]{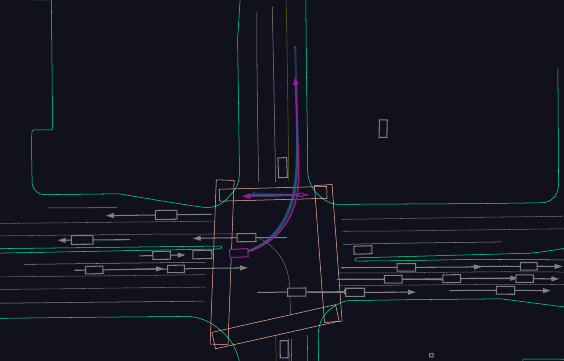}
    \includegraphics[width=1.0\textwidth,height=0.75\textwidth]{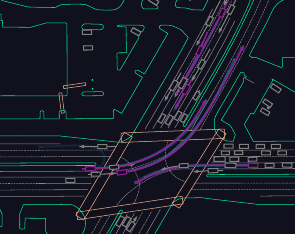}
    \includegraphics[width=1.0\textwidth,height=0.75\textwidth]{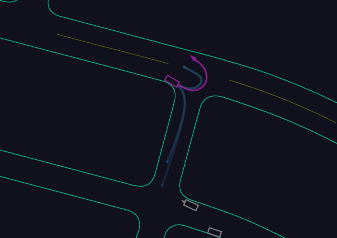}
    \includegraphics[width=1.0\textwidth,height=0.75\textwidth]{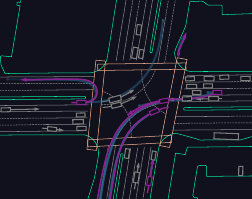}
    \caption{\#Mode@Training=3, \#Mode@Inference=3}
  \end{subfigure}
  \begin{subfigure}{0.33\linewidth}
    \includegraphics[width=1.0\textwidth,height=0.75\textwidth]{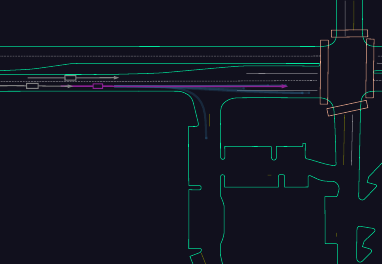}
    \includegraphics[width=1.0\textwidth,height=0.75\textwidth]{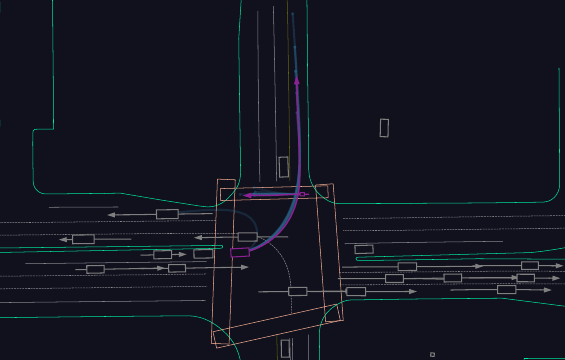}
    \includegraphics[width=1.0\textwidth,height=0.75\textwidth]{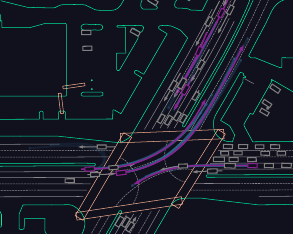}
    \includegraphics[width=1.0\textwidth,height=0.75\textwidth]{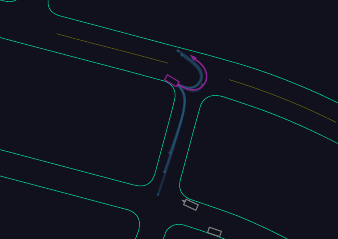}
    \includegraphics[width=1.0\textwidth,height=0.75\textwidth]{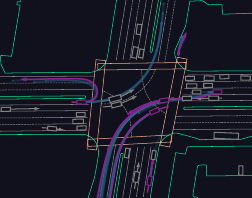}
    \caption{\#Mode@Training=6, \#Mode@Inference=6}
  \end{subfigure}
  \begin{subfigure}{0.33\linewidth}
    \includegraphics[width=1.0\textwidth,height=0.75\textwidth]{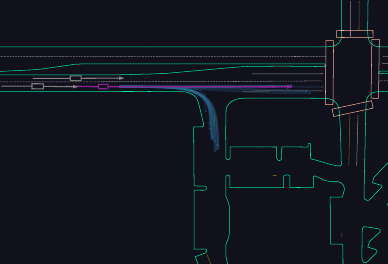}
    \includegraphics[width=1.0\textwidth,height=0.75\textwidth]{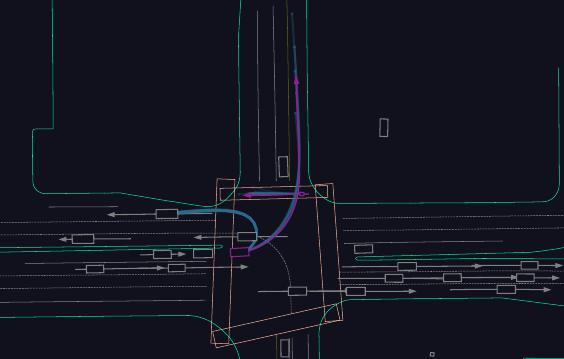}
    \includegraphics[width=1.0\textwidth,height=0.75\textwidth]{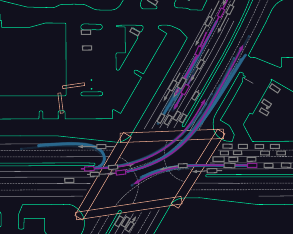}
    \includegraphics[width=1.0\textwidth,height=0.75\textwidth]{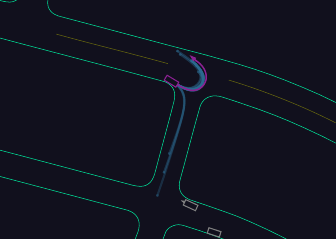}
    \includegraphics[width=1.0\textwidth,height=0.75\textwidth]{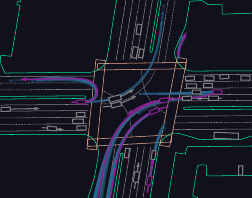}
    \caption{\#Mode@Training=6, \#Mode@Inference=24}
  \end{subfigure}
  \caption{Visualization on the WOMD. The agents in purple are predicted with blue trajectories, with the opacity indicating confidence.}
  \label{fig:more_qualitative}
\end{figure*}

\Cref{fig:more_qualitative} supplements the results in \cref{fig:qualitative} to demonstrate our approach's ability to produce representative trajectories and extrapolate more modes.

\end{document}